\documentclass[lettersize,journal]{IEEEtran}
\usepackage{amsmath,amsfonts}
\usepackage{algorithmic}
\usepackage{algorithm}
\usepackage{array}
\usepackage{textcomp}
\usepackage{stfloats}
\usepackage{url}
\usepackage{verbatim}
\usepackage{graphicx}
\usepackage{cite}
\usepackage{float}
\usepackage{multirow}
\usepackage{placeins}
\usepackage{listings}
\usepackage{tabularx}
\usepackage{xcolor}
\usepackage{lscape}
\usepackage{makecell}
\usepackage{subcaption}

\definecolor{c_comments}{RGB}{34, 139, 34}
\begin{document}

\title{SPILDL: A Scalable and Parallel Inductive Learner in Description Logic}

\author{Eyad Algahtani, King Saud University, Saudi Arabia}

\markboth{Journal of \LaTeX\ Class Files, December~2024}%
{Shell \MakeLowercase{\textit{et al.}}: A Sample Article Using IEEEtran.cls for IEEE Journals}


\maketitle

\begin{abstract}
We present SPILDL, a Scalable and Parallel Inductive Learner in Description Logic (DL). SPILDL is based on the DL-Learner (the state of the art in DL-based ILP learning). As a DL-based ILP learner, SPILDL targets the $\mathcal{ALCQI}^{\mathcal{(D)}}$ DL language, and can learn DL hypotheses expressed as disjunctions of conjunctions (using the $\sqcup$ operator). Moreover, SPILDL's hypothesis language also incorporates the use of string concrete roles (also known as string data properties in the Web Ontology Language, OWL); As a result, this incorporation of powerful DL constructs, enables SPILDL to learn powerful DL-based hypotheses for describing many real-world complex concepts.
SPILDL employs a hybrid parallel approach which combines both shared-memory and distributed-memory approaches, to accelerates ILP learning (for both hypothesis search and evaluation).
According to experimental results, SPILDL's parallel search improved performance by up to $\sim$27.3 folds (best case). For hypothesis evaluation, SPILDL improved evaluation performance through HT-HEDL (our multi-core CPU + multi-GPU hypothesis evaluation engine), by up to 38 folds (best case). By combining both parallel search and evaluation, SPILDL improved performance by up to $\sim$560 folds (best case). 
In terms of worst case scenario, SPILDL's parallel search doesn't provide consistent speedups on all datasets, and is highly dependent on the search space nature of the ILP dataset. For some datasets, increasing the number of parallel search threads result in reduced performance, similar or worse than baseline. Some ILP datasets benefit from parallel search, while others don't (or the performance gains are negligible). In terms of parallel evaluation, on small datasets, parallel evaluation provide similar or worse performance than baseline.
\end{abstract}


\section{Introduction}
\label{sec:Introduction}
\IEEEPARstart{I}{}nductive logic programming (ILP) is a Machine Learning (ML) technique which represents knowledge and models using logic-based representations. ILP as a machine learning technique, is used in several areas \cite{ilp_apps,ilp_bio,ilp_aero} for learning complex concepts from multi-relational data; this ability of directly learning from multi-relational data, gives ILP an advantage (in terms of model expressivity) when compared to other techniques such as ID3 and Naïve Bayes. A key advantage in ILP, is in the flexibility of its knowledge representation, that is, it is not tied to a particular logic representation. In fact, ILP has been used with other logic formalisms such as Description Logic (DL) \cite{dl_learner,dl_foil}; DL is typically used as the underlying knowledge representation for OWL (Web Ontology Language)\cite{owl_lang}, where DL also provide reasoning facilities on OWL ontologies. DL-based ILPs improve learning performance (and scalability) by reducing model expressivity into the midpoint between horn clauses and propositional logic (PL). Even though DL-based ILPs have reduced model expressivity (when compared to classical ILPs), yet they still retain a sufficient level of model expressivity to describe many complex real-world concepts. There are also other ML techniques capable of learning from multi-relational data such as artificial neural networks (ANN); however, such ML techniques provide black-box models which can not be interpreted by humans, whereas ILP algorithms learn white-box (human readable) models. 

Even though ILP algorithms are capable of learning complex human interpretable models, yet they suffer from poor scalability in terms of handling (or coping) with large learning data; because ILP algorithms are highly sequential in nature. Due to this poor scalability, ILP's potential real-world applications are limited to medium to small datasets. To address the scalability issue, several techniques were developed to improve the scalability of ILP algorithms such as reducing redundant computations through Query Packs\cite{query_packs}, outsourcing hypothesis evaluation (a key component in ILP algorithms) to database systems \cite{quick_foil}. Also, other developed approaches focused on improving ILP performance by parallelizing the hypothesis search\cite{parallel_ilp_search}. However, most of the performance improvement approaches in the ILP literature, focus mainly on improving the performance for classical ILPS -- i.e. ILPs that uses Horn clauses for knowledge and model representation. In terms of DL-based ILP literature, research works focus mainly on improving the hypothesis evaluation task for DL-based ILPs, such as through parallel DL reasoners\cite{konclude}. In other words, we believe that there are no parallel ILP approaches that accelerate hypothesis search for DL-based ILPs. 

Therefore, in this work we propose a parallel DL-based ILP learner that employ a set of novel parallel approaches that accelerate the hypothesis search for DL-based ILP learning. In our proposed parallel DL-based ILP learner (SPILDL, a scalable and parallel inductive learner in DL), we provide parallel hypothesis search approaches, that utilize shared-memory and also distributed-memory architectures, to accelerate the performance of DL-based ILPs. In terms of accelerating hypothesis evaluation, SPILDL outsource hypothesis evaluation to HT-HEDL\cite{ht_hedl} (our multi-device hypothesis evaluation engine); HT-HEDL aggregates the computing power of multi-core CPUs with multi-GPUs for high performance hypothesis evaluation in DL. SPILDL (our proposed work) builds and extends upon the DL-Learner's OCEL (OWL Class Expression Learner) algorithm, where the DL-Learner\cite{dl_learner} and its OCEL algorithm are regarded as the state of the art in the DL-based ILP literature. In the next section, we review the parallel and non-parallel ILP literature.

\section{Related work}
\label{sec:Related_work}
An ILP algorithm consists of three main procedures: hypothesis generation, hypothesis search and hypothesis evaluation; accelerating one or more of these steps will improve ILP performance. In terms of hypothesis generation, some existing techniques employ problem-related knowledge to restrict the hypothesis language to avoid generating invalid candidate hypotheses. Classical ILPs use mode declarations such as Progol\cite{progol} and Aleph\cite{aleph}. For DL-based ILPs, the DL-Learner employ class/role hierarchy and statistics about the knowledge base such as max role fillers and value boundaries for numeric concrete roles; when generating DL refinements. Other approaches improve hypothesis generation by detecting and avoiding the expansion of weakly-equal hypotheses (i.e. the same hypothesis though with different operands order); the DL-Learner employ such approach by enforcing a deterministic operand order on DL hypotheses. In addition, a maximum hypothesis length may be used to restrict the hypothesis language further.
In some ILP learning problems, it would be sufficient to use less expressive logic-based representations to learn an acceptable solution; in other words, using expressive logic representations is not always recommended -- especially if the ILP task at hand doesn’t require a high level of hypothesis expressivity. Less expressive logic representations generate simpler hypotheses with faster learning speed.

In terms of hypothesis search, there are two existing approaches. In the first approach, techniques were proposed to optimize the search process such as developing better scoring functions \cite{quick_foil}. In the second approach, techniques were developed which focuses on dedicating more computing power towards the search (and evaluation) process. Some approaches were developed to accelerate the Aleph ILP algorithm using Message Passing Interface (MPI)\cite{open_mpi} in \cite{mpi_ilp}. Some parallel ILP approaches accelerated the search using shared-memory environment \cite{shared_ilp_search_eval}, while others focused on distributed-memory environments\cite{dist_ilp_search}. In terms of hypothesis evaluation, some approaches focused on accelerating the hypothesis evaluation by using: concurrent logic programming languages\cite{Meissner_concurrent}, database systems such as Datalogs\cite{relational_ilp_gpu}, or parallel reasoners \cite{parallel_dl_reasoner}. Moreover, some developed approaches used GPUs to accelerate hypothesis evaluation\cite{parallel_dl_gpu,gpu_el_reasoner,gpu_rdf,gpu_sparql,gpu_rdf_opt}. Furthermore, there are also approaches that use dedicated hardware (FPGA-based) accelerators to improve ILP computations\cite{ilp_fpga,eyad_fpga}.

\begin{figure}[h]
\includegraphics[width=0.48\textwidth]{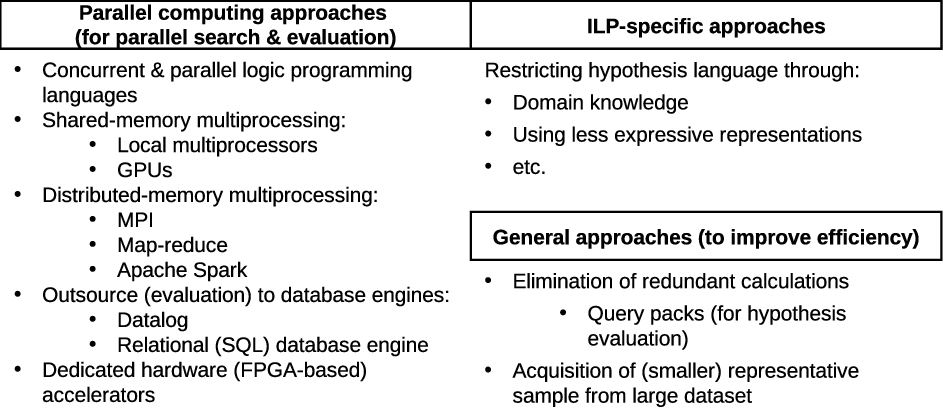}
\caption{A summary of parallel and non-parallel ILP literature.}
\label{fig:lit_summary}
\end{figure}

There are ILP acceleration approaches that use Big Data\cite{big_data} technologies (such as MapReduce\cite{map_reduce} and Apache Spark\cite{spark}) to accelerate ILP computations\cite{map_reduce_ilp,parallel_prolog,data_task_parallelism,mp_hthedl}. Although, one of the limitations (or challenges) with distributed parallel ILP approaches, is dealing with the search, coordination and communication overheads; the search overhead refers to the additional number of generated candidate hypotheses as opposed to the traditional (sequential) implementation. The communication overhead refers to the communication (network) cost between the distributed computers working on a single ILP problem; this overhead is small in shared-memory ILPs, though it is more pronounced (amplified) in distributed-memory ILPs. See Fig~\ref{fig:lit_summary} for a summary of the reviewed literature.

The reviewed approaches focused mostly on improving classical ILPs (horn clauses). We have observed rarity on improving DL-based ILPs through either parallel or non-parallel approaches. Therefore, in this article, we focus on improving DL-based ILPs through the use of parallel computing approaches to accelerate both of hypothesis search and hypothesis evaluation. In the next section, we describe SPILDL (our proposed approach) which combines existing non-parallel approaches (employed by the state of the art, i.e. DL-Learner) with our proposed parallel approach, for improving the performance of DL-based ILPs.

\section{SPILDL: Scalable and parallel inductive learning in DL}
\label{sec:SPILDL_sec}

In this section, we describe SPILDL, a Scalable and Parallel Inductive Learner in DL. SPILDL is a parallel ILP learner based on the DL-Learner. SPILDL aims to extends upon the DL-Learner by exploiting parallel computing capabilities of multi-core CPUs and multi-GPUs, to improve the performance of DL-based ILPs; in order to reduce ILP learning time and improve the scalability (i.e. coping with very large datasets). It is worth noting that some aspects of SPILDL in \cite{spildl_thesis}, are used in our previous works in \cite{ilp2018_paper,ilp2019_paper}.

SPILDL is based on the DL-Learner's OCEL algorithm, and follows OCEL's search procedure (including expanding hypotheses up to $he$), scoring function, and its full refinement operator; which also includes constructing hypotheses with the $\sqcup$ operator, e.g. $(C_1 \sqcap C_2) \sqcup (C_3 \sqcap C_4)$. In addition, SPILDL also increases the expressivity of learned hypotheses by incorporating string concrete roles (e.g. $injuryLevel="severe"$, discussed in next section), in addition to numeric concrete roles, when constructing candidate hypotheses; this incorporation of expressive DL constructs, results in learning much more expressive DL hypotheses suitable for wider range of DL-based inductive learning applications.

SPILDL targets the $\mathcal{ALCQI}^{\mathcal{(D)}}$ DL language. Similar to OCEL, SPILDL can also incorporate role hierarchy when generating refinements. In terms of generating hypotheses with string roles, SPILDL handles string roles (e.g. $stringProperty="value"$), similar to how OCEL treats boolean concrete roles; i.e. for each string role, SPILDL adds to the $M_B$ set, all possible values for each string role. For example, given the string role $sr$ which has the possible values (according to ABox assertions): ${"val1", "val2", "val3"}$; SPILDL will add to $M_B$ set: $sr="val1"$, $sr="val2"$ and $sr="val3"$. We believe new versions of DL-Learner's algorithms, may already support constructing hypotheses with string roles using similar approach; however, in the context of this work, we focus on the scalable (and performance) aspect of learning hypotheses with string roles. In terms of SPILDL's hypothesis language, SPILDL builds hypotheses using the following DL constructors:

\begin{itemize}
\item atomic concepts and negated atomic concepts
\item existential, universal and role cardinality restrictions
\item boolean ($b=true$, $b=false$), float ($d \geq v$, $d \leq v$) and string ($sr=strVal$) role restrictions
\item conjunctions of above simple/complex concepts
\subitem including conjunctions of disjunctions.
\item disjunctions of above simple/complex concepts 
\subitem including disjunctions of conjunctions
\end{itemize}

Next, we describe SPILDL's parallel computing approaches for accelerating ILP learning.

\subsection{High-performance ILP learning}
\label{sec:SPILDL_parallel_learning}
\begin{figure}[h]
\includegraphics[width=0.45\textwidth]{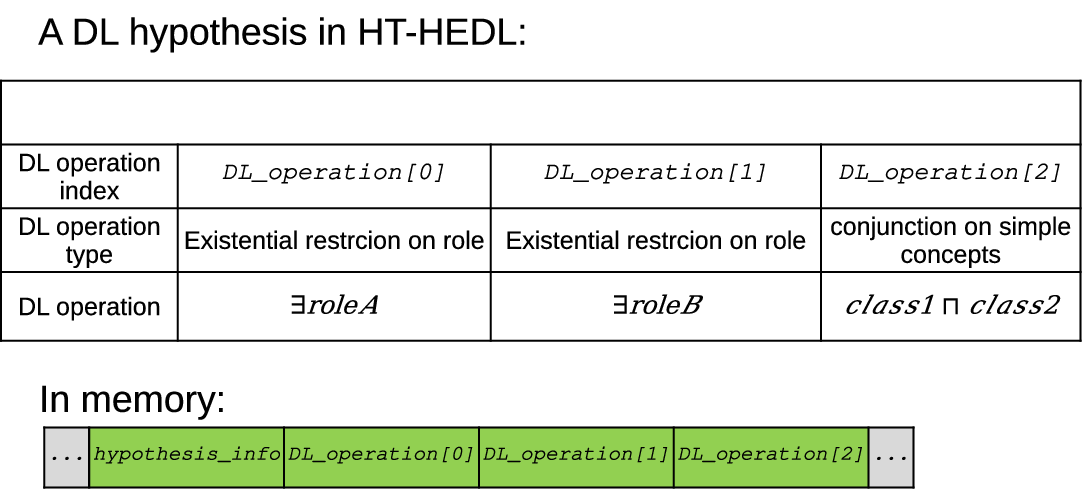}
\caption{HT-HEDL's DL hypothesis representation\cite{ht_hedl}.}
\label{fig:ht-hedl_hyp}    
\end{figure}

\begin{figure}[h]
\includegraphics[width=0.45\textwidth]{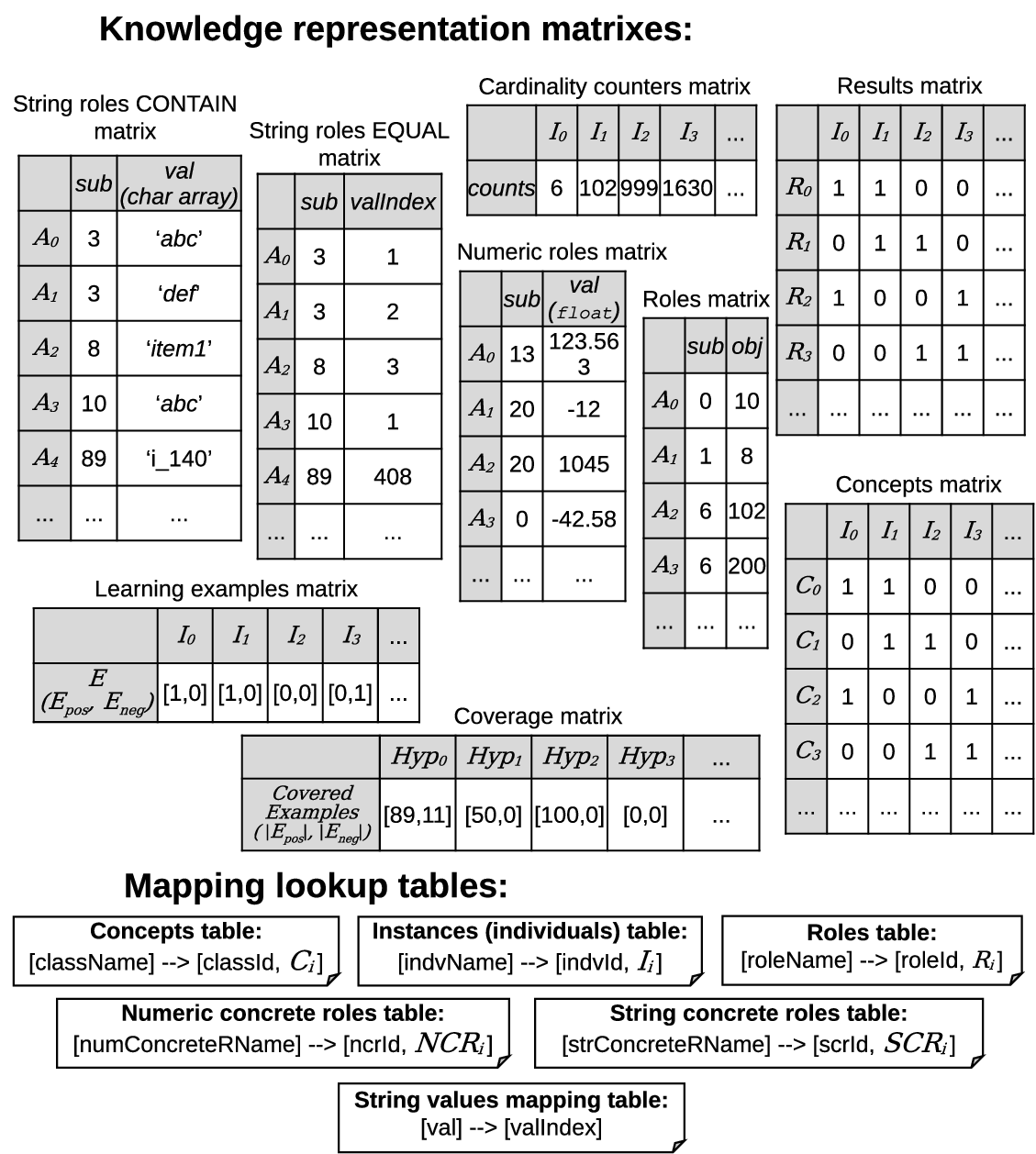}
\caption{HT-HEDL's knowledge representation\cite{ht_hedl}.}
\label{fig:ht-hedl_kb}    
\end{figure}

SPILDL aggregates the computing power of local and network-linked heterogeneous processors (both GPUs and CPUs), to accelerate ILP learning. In terms of hypothesis search, SPILDL employs parallel hypothesis search which is performed by both local and networked (cluster-based) multi-core processors. In terms of hypothesis evaluation, SPILDL outsources hypothesis evaluation task to HT-HEDL, our novel high-performance multi device hypothesis evaluation engine; HT-HEDL aggregates the computing power of multi-CPUs (with their vector instructions) and multi-GPUs to accelerate hypothesis evaluation at the level of a single hypothesis and at the level of multiple hypotheses (a batch of hypotheses). See Fig.~\ref{fig:ht-hedl_kb} and Fig.~\ref{fig:ht-hedl_hyp} for HT-HEDL's knowledge and DL hypothesis representation, respectively; HT-HEDL's hypothesis and knowledge representations are optimized for high performance hypothesis evaluation, and with minimal serialization/deserialization overheads (for remote evaluation). The symbols in Fig.~\ref{fig:ht-hedl_kb} are described in Table~\ref{tab:sym_tab}.

 \FloatBarrier
 
 \FloatBarrier

\begin{table}[!htb]
\caption{Description for HT-HEDL's knowledge representation symbols\cite{ht_hedl}.}
\label{tab:sym_tab}
\begin{tabular}{m{1.05cm}|m{2.6cm}|m{3.9cm}}
\hline
\textbf{Symbol} & \textbf{Related Matrixes} & \textbf{Symbol description} \\\hline
$C_i$ & Concepts (classes) matrix & represent a particular concept \\\hline
$I_i$ &\begin{itemize}
\item Concepts matrix
\item Results matrix
\item Cardinality counters matrix
\item Learning examples matrix
\end{itemize} & represent a particular individual. \\\hline
$R_i$ & Results matrix & represent an intermediate result for a particular DL operation. \\\hline
$A_i$, $sub$ &\begin{itemize}
\item Roles matrix
\item Numeric roles matrix
\item String roles EQUAL matrix
\item String roles CONTAIN matrix
\end{itemize} & $A_i$:represent a single role or a concrete (string/numeric) role assertion.\newline $sub$: refers to the subject of an assertion, which is an individual ($I_i$).\\\hline
$obj$ & Roles matrix & refers to the object of an assertion, which is an individual ($I_i$).\\\hline
$val$ &\begin{itemize}
\item Numeric roles matrix
\item String roles CONTAIN matrix
\end{itemize} & refers to the value of a given concrete role assertion.\\\hline
$valIndex$ & String roles EQUAL matrix & a numeric (integer) mapping of a string assertion's value from String values mapping table (in Fig.~\ref{fig:ht-hedl_kb}).\\\hline
$E(E_{pos},$\newline$E_{neg})$ & Learning examples matrix & $E_{pos}$: is a binary value that indicate whether the given individual $I_i$ is a positive example or not.\newline $E_{neg}$: indicate whether the given individual $I_i$ is a negative example or not.\\\hline
$Hyp_i$\newline$(|E_{pos}|,$\newline$|E_{neg}|)$ & Coverage matrix & for a given hypothesis $Hyp_i$, $|E_{pos}|$ represent the number of covered positive examples, and $|E_{neg}|$ represent the number of covered negative examples. \\\hline
\end{tabular}
\end{table}

SPILDL has two modes for ILP learning, shared-memory and cluster-based learning.
\subsection{Shared-memory learning}
\label{sec:shared_memory}

SPILDL accelerates the hypotheses search for OCEL, by employing parallel (multi-threaded) beam search. SPILDL modifies OCEL's search algorithm in order to exploit parallel computing performance from multi-threaded processors. Even though SPILDL modifies OCEL search algorithm, though it still retains OCEL's ILP learning characteristics; In other words, when SPILDL conducts hypothesis search using only a single thread (i.e. sequential search), it should possess the same ILP learning characteristics as OCEL. 
The SPILDL algorithm is defined in Algorithm~\ref{alg:SPILDL_algorithm}.

\begin{algorithm}[h]
 \caption{SPILDL algorithm for shared-memory learning \cite{spildl_thesis}}
 \label{alg:SPILDL_algorithm}    
 \begin{lstlisting}[numbers=none]
//Input: search tree array ST, 
//the redundancy hash table RHT, beam width BW, 
//number of final solutions LIMIT
//Output: list of learned hypotheses LH

while ST has expandable nodes
  N = extractBestNodes(ST,BW)
  
  for every node n in N
   //expand node, sort refinements, check redundancy
   //and add local CPU's closed list
   fork expandSingleNode(n) 
  end for
  
  //wait for all threads to finsh
  join threads 
  
  //parallel refinements reduction
  FinalRefs = computeFinalNonRedundantNodes(N)
  Call addRefinmntsToRedndncyTble(FinalRefs,RHT)

  GoodRefs = computeEvaluationResults(FinalRefs)
  
  Call addRefinementsToSearchTree(GoodRefs,ST)
  
  Call parallelSortSearchTree(ST)
end while

//return the best limit hypotheses
LH = extractBestNodes(ST,LIMIT)
return LH
\end{lstlisting}
\end{algorithm}

In Algorithm~\ref{alg:SPILDL_algorithm}, SPILDL maintains an open list ($ST$) and a closed list ($RHT$). the reason for using an array for the open list (instead of, for example, a priority queue) is for performance reasons, that is, SPILDL may add large number of hypotheses to the open list at once, with which an array is much more suitable than a sorted hypothesis container; we use parallel sorting (in line 21) on the open list array in order to improve performance, and to avoid potential performance bottleneck due to using sequential (non-parallel) sorting on the open list.

\begin{figure}[h]
\includegraphics[width=0.48\textwidth]{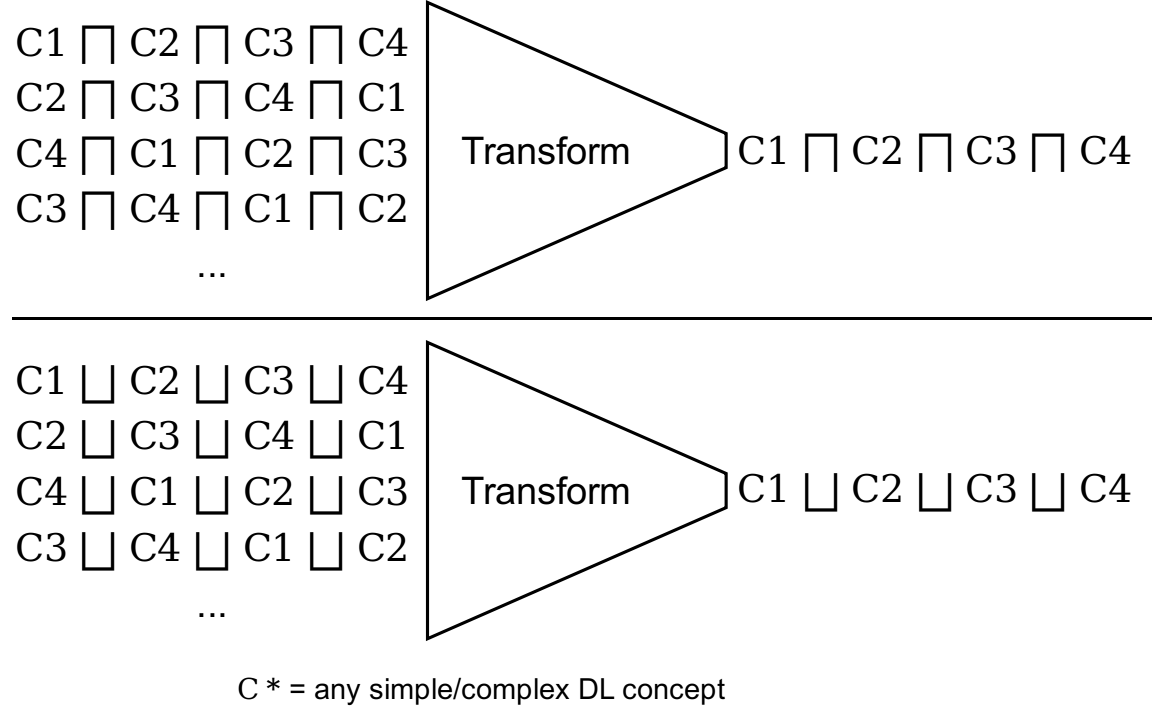}
\caption{Deterministic ordering of hypotheses' conjunction/disjunction operands.}
\label{fig:redCheck}    
\end{figure}

The learning starts by getting the best $BW$ nodes from the open list, where $BW$ is set to the number of CPU cores. After the best $BW$ nodes (with highest OCEL score) are extracted, the nodes are then expanded (refined) in parallel by each CPU core. Each CPU core generates refinements for its assigned node (hypothesis), sort each generated refinement's operands (see Fig.~\ref{fig:redCheck}); sorting refinement's operand, means the reordering of conjunction/disjunction operands (sub DL concepts) to a deterministic order, which will then be used to detect and eliminate weakly-equal hypotheses. When a generated refinement is reordered, a hash value is computed and stored within the refinement for more efficient (faster) redundancy checking. After each CPU core, finished generating its refinements and sorting (their operands), it will then check its refinements' redundancy against the main closed list ($RHT$).

\FloatBarrier
 
\begin{figure}[h]
\includegraphics[width=0.43\textwidth]{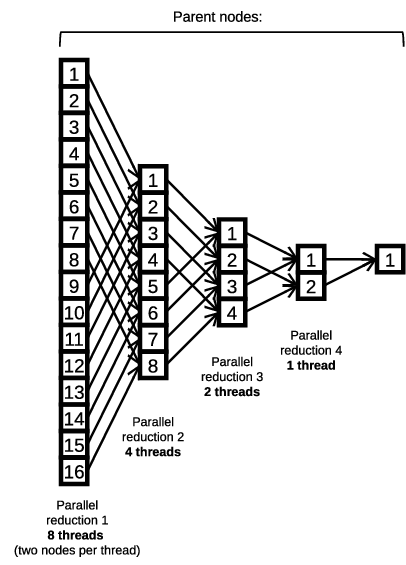}
\caption{Parallel multi-stage reduction for redundancy checking.}
\label{fig:parallel_red_check}   
\end{figure}

Once all parallel CPU cores complete their hypothesis expansion task (in line 7), a parallel reduction is performed for checking the redundancy of generated refinements (by all CPU cores), against each other to get the final list of non-redundant refinements; see Fig.~\ref{fig:parallel_red_check} for the parallel redundancy check. Each CPU core has its local closed list which contains the hash values for its generated refinements (by the step in line 7), these local closed lists are used by the parallel reduction to improve efficiency, by reducing the cost of checking a single refinement to approximately O(1); when CPU core\textsubscript{i}'s refinements are checked against CPU core\textsubscript{j}'s.

\FloatBarrier

Once the parallel reduction is complete, it will result in the final list of unique (non-redundant) refinements; these refinements will then be evaluated using HT-HEDL (in line 17). Next, the evaluated refinements with non-weak OCEL scores, will be added into the open list, and then a parallel sorting is performed on the open list. 

In the next section, we describe how SPILDL learns using clustered (networked) computers in order to parallelize the hypothesis search beyond local (shared-memory) environment.

\subsection{Cluster-based learning}
We have described SPILDL's shared-memory learning in the previous section, which uses the shared-memory parallel model. Other types of parallel computing models exist as well such as the distributed-memory model, which accelerates computations through remote (networked) processors. SPILDL's cluster-based learning combines the shared and distributed -memory models into a hybrid model, in order to maximize performance advantages of each model while minimizing their limitations. In this hybrid model, SPILDL combines both local (multi-threaded) and remote processors to conduct the parallel hypothesis search. For hypothesis evaluation, each machine in the cluster will evaluate its generated refinements using HT-HEDL. 

SPILDL's cluster-based learning follows the master-worker model. In this setup, the master node (machine) manages, coordinates, and assigns work to worker nodes (which also includes collecting their computing results).

SPILDL's cluster-based learning has 4 phases: discovery, probing, learning and termination phases. In the discovery phase, the master identifies available worker nodes in the network. In the probing phase, the master gauges (or probes) the hypothesis search and evaluation capabilities through a dummy load and then measure its execution time (similar to HT-HEDL's approach). In the learning phase, the master now will conduct the ILP learning assign hypothesis search and evaluation tasks, to each worker based on its computing capabilities. In the next sections, we describe each phase in detail.

\subsubsection{The discovery phase}
\begin{figure}[h]
\includegraphics[width=0.49\textwidth]{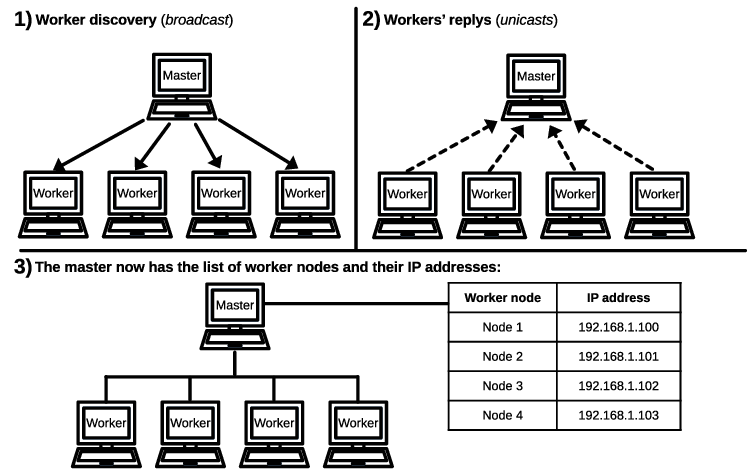}
\caption{An overview of the discovery phase.}
\label{fig:3}    
\end{figure}

In the discovery step, the master node broadcasts a message (using a UDP broadcast packet) to all machines (potential workers) in the network cluster, all machines will then replay back with a message to indicate their presence. The master will then establish a TCP connection for each worker (that replied to the broadcast message), and binds it to a dedicated CPU thread in the master's machine; the master node will now have multiple concurrent TCP connections, where each worker has a dedicated TCP connection to the master node. The reason for having multiple concurrent TCP connections, is to maximize the utilization of available network bandwidth (to improve efficiency). Even though these concurrent connections are not actually sending and receiving data in parallel, because they share the same communication channel. However, combining concurrent TCP connections with multi-threading (i.e. each parallel thread will send data through its TCP connection), will saturate the shared communication channel, thus improving the efficiency of network bandwidth.
See Fig.~\ref{fig:3} for an overview of the discovery step.
 
\FloatBarrier

\subsubsection{The probing phase}

\begin{figure}[h]
\includegraphics[width=0.48\textwidth]{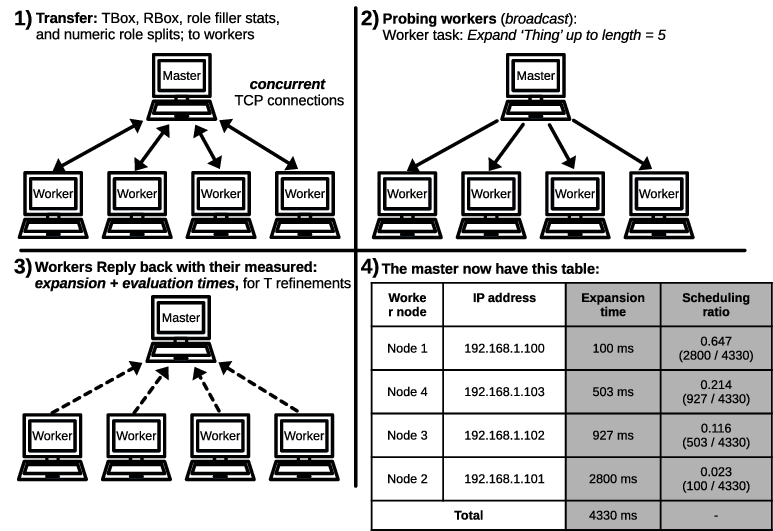}
\caption{An overview of the probing phase.}
\label{fig:4}    
\end{figure}

In the probing phase, the master will first send (through its TCP connections): TBox, RBox and ABox (in binary forms) to each worker. After all workers have received a copy of the knowledge base, the master will then issue to all workers to expand the top concept ($\top$, also known as 'owl:thing') with horizontal expansion ($he=5$). After that, each worker will evaluate its generated refinements through its local HT-HEDL evaluation; the worker will then measure the execution time for the top concept expansion + its refinements evaluation (using HT-HEDL). After that, each worker will reply back to the master with their CPU cores count and the measured execution time (for expansion + evaluation). The reason for expanding the $\top$ concept and measuring its execution time, is to evaluate hypothesis expansion (or refinement generation) capabilities for each worker; which will then be used to assign appropriate workload sizes for each machine. Once all workers reply back to the master with their probing results, the master machine will count the total number of CPU cores for all workers, and then use it to calculate the scheduling ratio for each worker. See Fig.~\ref{fig:4} for an overview of the probing step.

\FloatBarrier

\subsubsection{The learning phase}
\begin{figure}[h]
     \centering
     \begin{subfigure}[b]{0.45\textwidth}
         \centering
\includegraphics[width=0.8\textwidth]{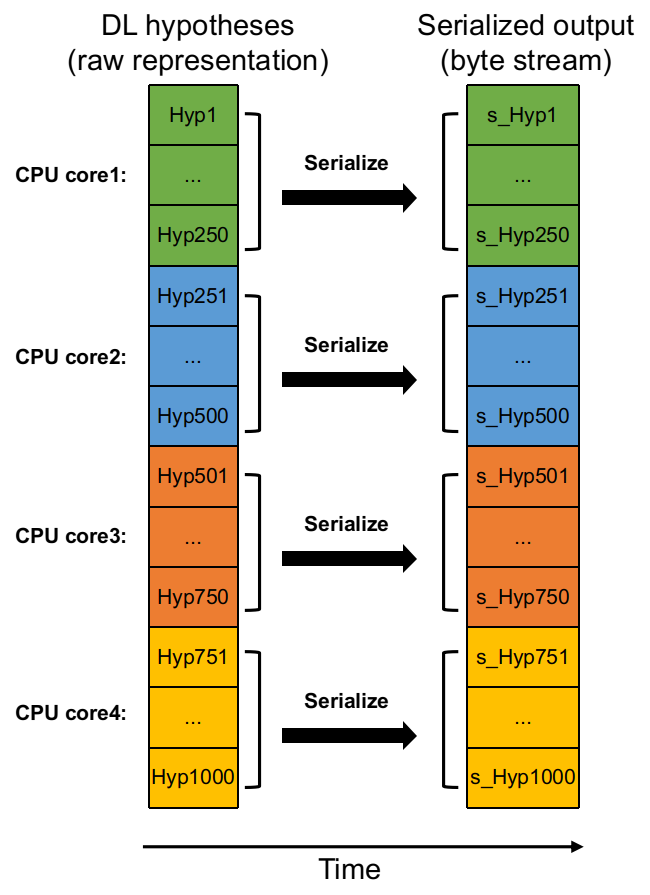}
         \caption{Parallel serialization of \\DL hypotheses}
     \end{subfigure}
     \hfill
     \begin{subfigure}[b]{0.45\textwidth}
         \centering
\includegraphics[width=0.8\textwidth]{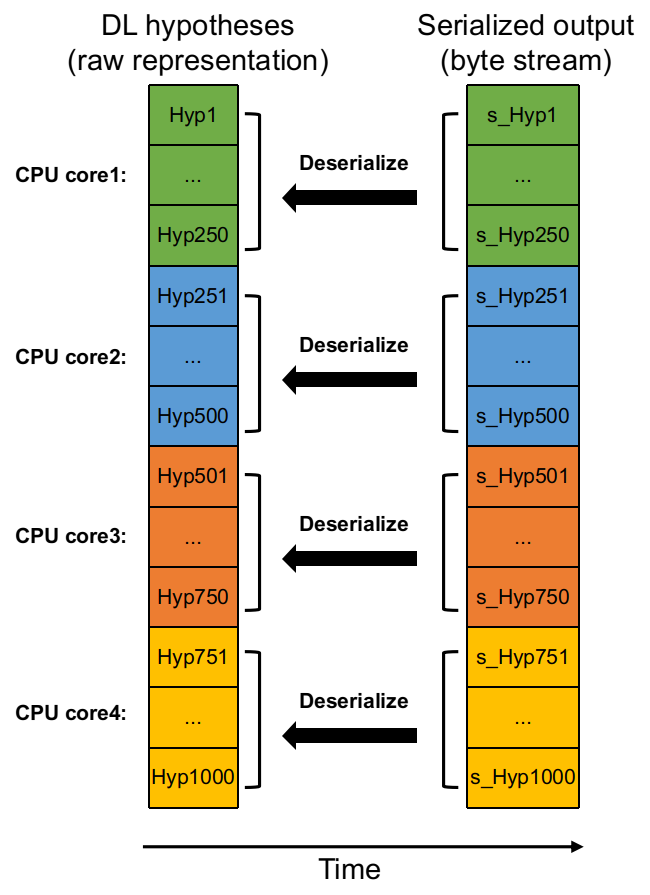}
         \caption{Parallel deserialization of DL \\hypotheses}
     \end{subfigure}
        \caption{SPILDL's multi-threaded hypothesis serialization/deserialization using 4 CPU cores.}
        \label{fig:parallel_serial_deserial}
\end{figure}

\begin{figure*}[h]
\includegraphics[width=0.9\textwidth]{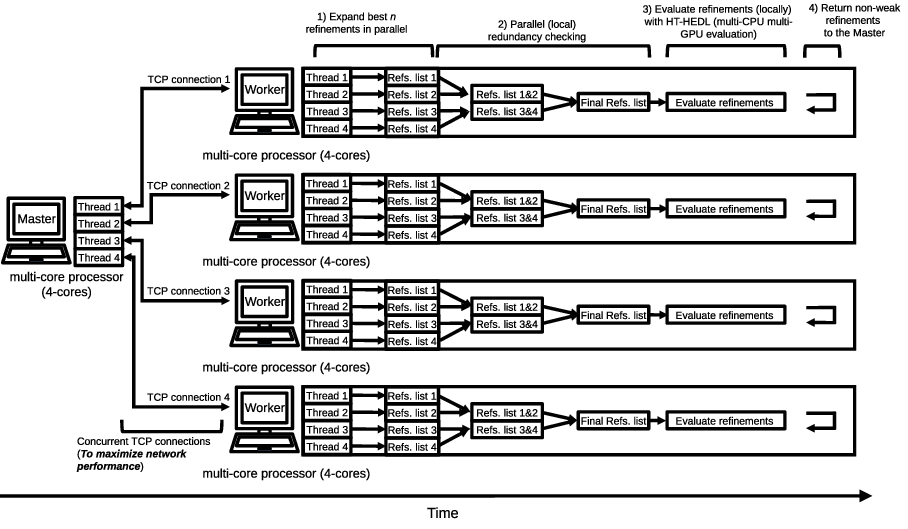}
\caption{An overview of SPILDL's cluster-based learning -- learning phase.}
\label{fig:SPILDL_cluster}
\end{figure*}

In the learning phase, similar to shared-memory learning, SPILDL takes the best $n$ nodes in the open list, expands them and evaluate their generated refinements in parallel. However, in cluster-based learning, SPILDL incorporates the multi-core CPUs of all workers in the network.
A major challenge in incorporating networked processors to improve computing performance, is the communication overheads, i.e. the network and serialization/deserialization overheads. We have implemented some measures to improve communication performance, such as several concurrent TCP communication channels between the master and workers. In terms of serialization/deserialization, we serialize and deserialize DL hypotheses (represented in HT-HEDL representation) in parallel, to minimize serialization/deserialization overhead. see Fig.~\ref{fig:parallel_serial_deserial} for SPILDL's parallel (multi-threaded) serialization/deserialization of hypotheses. Also see Fig.~\ref{fig:SPILDL_cluster} for SPILDL's ILP learning (in a cluster environment).
In Fig.~\ref{fig:SPILDL_cluster}, the master node maintains the open list and the main closed list, and each worker has its local closed list; the ILP learning starts in the master node. First, the master gets the best $n$ hypotheses from the open list, where $n$ is the total number of CPU cores across all workers (i.e. $n=16$). After that, from the best $n$ nodes, each CPU thread in the master will get (in parallel) the best $wn$ hypotheses for each worker, where $wn$ is the number of CPU cores in the worker's machine. Then, each master thread will serialize its $wn$ hypotheses (from their raw HT-HEDL's representation to byte stream representation), and then send it to its corresponding worker. 
 \FloatBarrier

\FloatBarrier

At this point, each worker will proceed similarly to shared-memory learning (discussed previously). However, the worker will receive its $wn$ hypotheses in a raw byte stream (from its TCP connection to the master); The worker will deserialize the received hypotheses (from bytes stream representation to HT-HEDL's representation). Once the hypotheses are in HT-HEDL's form, the worker will expand the hypotheses in parallel through multi-threading, then redundancy checking (through parallel reduction) until the final list of non-redundant refinements are produced. After that, the worker will evaluate its final list of non-redundant refinements using HT-HEDL (combining all CPUs and GPUs in the worker's machine). The worker's evaluated refinements which has non-weak scores are then serialized in parallel through multi-threading by the worker's multi-core processor, and then sent to the master. The master will receive the serialized hypotheses from workers. Once the master receives hypotheses from all workers, it will then deserialize them in parallel, and then will apply parallel redundancy checking against workers' refinements (and against the master's main closed list). The resulting non-redundant and non-weak refinements are then added to the master's open list.

\subsubsection{The termination phase}
in the termination phase, if one of the workers reports back to the master with a hypothesis that satisfies the user-defined conditions (e.g. minimal acceptable accuracy, learning timeout), SPILDL will then terminates the learning by broadcasting to all worker nodes to stop learning. After that, workers will send their best discovered hypotheses to the master.

\section{Implementation}
We implement SPILDL in C/C++ language, and we use the OpenMP API\cite{openmp} for multi-threading in both shared and clustered -environments. For hypothesis evaluation, HT-HEDL is implemented in C/C++, and Nvidia’s CUDA API\cite{cuda} is used for HT-HEDL's GPU-based evaluation. For CPU-based evaluation, HT-HEDL uses the SSE instruction set (available in many x86 and x64 CPU architectures) in combination with OpenMP, to facilitate CPU-based vectorized multi-threaded evaluation.
For cluster-based learning, the master and worker machines are connected using TCP/IP networks; the UDP protocol is used for broadcasting messages from the master to worker nodes, and TCP protocol for the set of dedicated (concurrent) connections between the master and each worker. For materializing the TBox and RBox, SPILDL uses the Hermit\cite{hermit} DL reasoner. To aid SPILDL's implementation, we use the DL-Learner's Java implementation in~\cite{dl_learner_src} as a supplementary material, to determine key algorithmic details for the OCEL algorithm.

\section{Experiments and discussion}
\label{sec:exp_results}
In this section, we provide experimental results to evaluate SPILDL's: shared-memory and cluster based learning. For description of the hardware used in the experiments, see Table~\ref{tab:m_spec}. The reported execution times are in milliseconds for all experiments. We use the DL-version of classical ILP datasets (Michalski's trains, Moral reasoner and Carcinogenesis) from the DL-Learner's repository in~\cite{dl_learner_src}.

\begin{table}
\caption{The machines' setup for SPILDL experiments.}
\label{tab:m_spec}
\begin{tabular}{c|p{3cm}|p{2.8cm}}
\hline
\textbf{Machines} & Machine 1 (M\textsubscript{1}) & Machine 2 (M\textsubscript{2}) \\\hline
\textbf{CPU(s)} & CPU\textsubscript{1}: AMD 5950x (16 core CPU) & AMD 3750H (8 core CPU)\\\hline
\textbf{Main memory} & 32 GB & 16 GB\\\hline
\textbf{\multirow{2}{*}{GPU(s)}} & GPU\textsubscript{1}: Nvidia GTX 1070 & Integrated on-chip GPU\\ 
& GPU\textsubscript{2}: Nvidia GTX 1060 & \\
& GPU\textsubscript{3}: Nvidia GTX 970  & \\\hline
\end{tabular}
\end{table}

In terms of ILP learning, we evaluate SPILDL learning using 6 datasets which can be seen in Table~\ref{tab:datasets_tab}. The 6 datasets vary in size in terms of ABox (which affects hypothesis evaluation), and TBox \& RBox (which affects the size and nature of the search space); these datasets will provide a representative measure, which reflects SPILDL's learning performance against real-world applications.

\begin{table}
\caption{The datasets used in the experiments.}
\label{tab:datasets_tab}
\begin{tabularx}{\columnwidth}{p{1cm}|p{1cm}|p{1cm}|p{1cm}|p{1cm}|p{1cm}}
\hline
 & Michalski Trains & Moral reasoner & Carcino-genesis & IMDB\textsuperscript{1} and IMDB (string version)\textsuperscript{2} & Dunn-humby retail\textsuperscript{3} \\ \hline
\#Classes & 10 & 44 & 142 & 27 & 81 \\ \hline
\#Roles & 5 & 0 & 4 & 1 & 1 \\ \hline
\#Concrete roles & 0 & 0 & 15 & 1 & 2 \\ \hline
\#Indivs & 50 & 202 & 22372 & 1224835 & 94838 \\ \hline
\multicolumn{6}{c}{-}   \\ \hline
\#Class asserts & 113 & 4646 & 22372 & 2437672 & 191181 \\ \hline
\#Role asserts & 149 & 0 & 40666 & 3431489 & 704275 \\ \hline
\#concrete. role asserts & 0 & 0 & 11185 & 388269 & 1602 \\ \hline
\multicolumn{6}{c}{-}   \\ \hline
\#positive examples & 5 & 102 & 182 & 139864 & 142 \\ \hline
\#negative examples & 5 & 100 & 155 & 677854 & 2358 \\ \hline

\multicolumn{6}{l}{\scriptsize{\textsuperscript{1}: Constructed from the dataset at https://relational.fit.cvut.cz/dataset/IMDb}} \\
\multicolumn{6}{l}{\scriptsize{\textsuperscript{2}: IMDB (string version) is a modified version of the constructed IMDB\textsuperscript{1} dataset}} \\
\multicolumn{6}{l}{\scriptsize{\textsuperscript{3}: Constructed from the dataset at https://www.kaggle.com/frtgnn/}} \\
\multicolumn{6}{l}{\scriptsize{dunnhumby-the-complete-journey}} \\

\end{tabularx}
\end{table}
\FloatBarrier

\subsection{The experiments for Shared-memory learning}
We start evaluating SPILDL by studying the effect of different evaluation techniques (CPU, GPU, GPU + CPU, etc.) on learning performance. For all the experiments for SPILDL's shared-memory learning, we use machine 1 (M\textsubscript{1}), since it has the most computing capabilities as opposed to machine 2 (M\textsubscript{2}); we also use GPU\textsubscript{1} for single GPU evaluation in M\textsubscript{1}, because it is M\textsubscript{1}'s most powerful GPU.

See Table~\ref{tab:shared_memory_learning_diff_eval} for the experimental results on shared-memory learning.
\begin{table}[!htb]
\caption{The experimental results for SPILDL's Shared-memory learning.}
\label{tab:shared_memory_learning_diff_eval}
\begin{tabularx}{\columnwidth}{p{1.5cm}|c|c|X|X|X|X}
\hline
 \multicolumn{7}{c}{\textbf{Using sequential scalar CPU evaluation (baseline)}} \\ \hline
 \multirow{2}{*}{\textbf{Dataset}} & \multicolumn{6}{c}{\textbf{Parallel search threads}} \\ \cline{2-7}
 & \textbf{1 (baseline)} & \textbf{2} & \textbf{4} & \textbf{8} & \textbf{16} & \textbf{32}\\ \hline
Michalski trains & 126 & 71 & 9 & 6 & 31 & 30\\ \hline
Moral & 1544 & 1355 & 1725 & 897 & 1087 & 1572\\ \hline
Carcino-genesis & 1791 & 1387 & 1190 & 10464 & 6690 & 3364\\ \hline
IMDB & 10608 & 6622 & 1302 & 1657 & 5138 & 8772\\ \hline
Dunn-humby Retail & 125032 & 111296 & 103582 & 100352 & 98058 & 96617\\ \hline
IMDB (string version) & 109117 & 69635 & 4002 & 76052 & 40195 & 49911\\ \hline

\multicolumn{7}{c}{\textbf{Using pure (scalar) multi-threaded CPU evaluation }} \\ \hline
 \multirow{2}{*}{\textbf{Dataset}} & \multicolumn{6}{c}{\textbf{Parallel search threads}} \\ \cline{2-7}
 & \textbf{1} & \textbf{2} & \textbf{4} & \textbf{8} & \textbf{16} & \textbf{32}\\ \hline
Michalski trains & 148 & 90 & 11 & 7 & 36 & 39\\ \hline
Moral & 1609 & 1378 & 1775 & 927 & 1081 & 1636\\ \hline
Carcino-genesis & 1511 & 1070 & 872 & 9364 & 5790 & 2625\\ \hline
IMDB & 1861 & 1083 & 279 & 262 & 935 & 1374\\ \hline
Dunn-humby Retail & 53576 & 39449 & 32368 & 28608 & 26572 & 25598\\ \hline
IMDB (string version) & 26371 & 15719 & 993 & 13098 & 10598 & 12145\\ \hline

 \multicolumn{7}{c}{\textbf{Using vectorized multi-threaded CPU evaluation}} \\ \hline
 \multirow{2}{*}{\textbf{Dataset}} & \multicolumn{6}{c}{\textbf{Parallel search threads}} \\ \cline{2-7}
 & \textbf{1} & \textbf{2} & \textbf{4} & \textbf{8} & \textbf{16} & \textbf{32}\\ \hline
Michalski trains & 148 & 89 & 11 & 7 & 35 & 35\\ \hline
Moral & 1518 & 1325 & 1717 & 907 & 1073 & 1605\\ \hline
Carcino-genesis & 1428 & 1008 & 808 & 9401 & 5554 & 2475\\ \hline
IMDB & 704 & 395 & 95 & 114 & 375 & 596\\ \hline
Dunn-humby Retail & 44887 & 31012 & 23917 & 20120 & 18006 & 16730\\ \hline
IMDB (string version) & 10706 & 7075 & 370 & 8572 & 3515 & 4140\\ \hline

 \multicolumn{7}{c}{\textbf{Using single GPU evaluation (GPU\textsubscript{1}, the fastest GPU available)}} \\ \hline
 \multirow{2}{*}{\textbf{Dataset}} & \multicolumn{6}{c}{\textbf{Parallel search threads}} \\ \cline{2-7}
 & \textbf{1} & \textbf{2} & \textbf{4} & \textbf{8} & \textbf{16} & \textbf{32}\\ \hline
Michalski trains & 154 & 101 & 12 & 8 & 40 & 46\\ \hline
Moral & 1492 & 1306 & 1680 & 905 & 1057 & 1607\\ \hline
Carcino-genesis & 1362 & 968 & 774 & 9294 & 5495 & 2423\\ \hline
IMDB & 472 & 268 & 60 & 82 & 214 & 364\\ \hline
Dunn-humby Retail & 48900 & 35315 & 28236 & 24423 & 22419 & 21174\\ \hline
IMDB (string version) & 8139 & 5921 & 256 & 7111 & 2004 & 2488\\ \hline

 \multicolumn{7}{c}{\textbf{Using (HT-HEDL's) multi-device evaluation (CPU\textsubscript{1}+GPU\textsubscript{1-3})}} \\ \hline
 \multirow{2}{*}{\textbf{Dataset}} & \multicolumn{6}{c}{\textbf{Parallel search threads}} \\ \cline{2-7}
 & \textbf{1} & \textbf{2} & \textbf{4} & \textbf{8} & \textbf{16} & \textbf{32}\\ \hline
Michalski trains & 158 & 97 & 12 & 8 & 39 & 47\\ \hline
Moral & 1542 & 1314 & 1678 & 908 & 1071 & 1596\\ \hline
Carcino-genesis & 1386 & 955 & 769 & 9231 & 5495 & 2387\\ \hline
IMDB & 279 & 195 & 41 & 44 & 112 & 210\\ \hline
Dunn-humby Retail & 50239 & 36695 & 29576 & 25612 & 23993 & 22938\\ \hline
IMDB (string version) & 8966 & 6236 & 195 & 6721 & 2484 & 3076\\ \hline
\end{tabularx}
\end{table}
\FloatBarrier

\begin{figure}[h]
\includegraphics[width=0.45\textwidth]{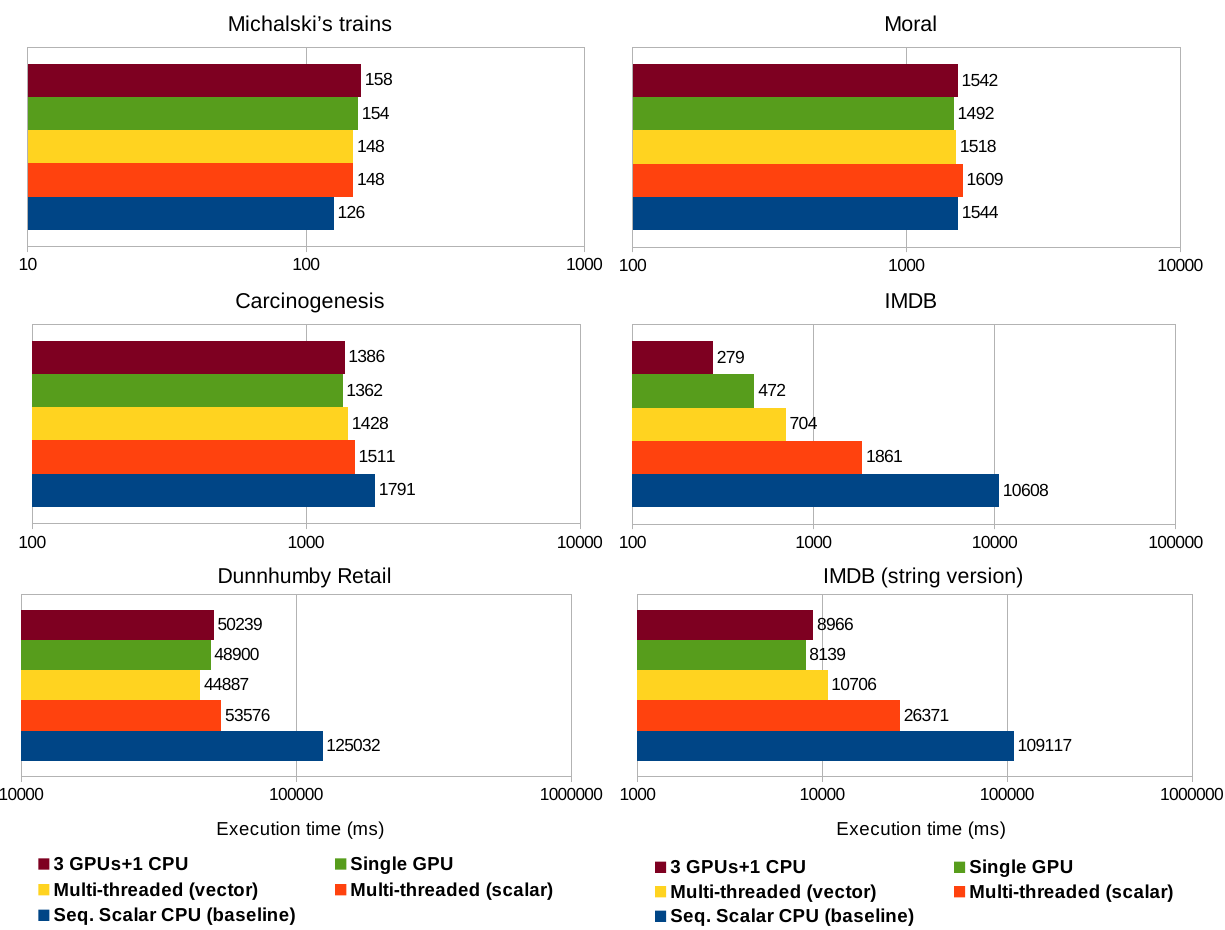}
\caption{The effect of different evaluation methods on learning performance (with seq. search).}
\label{fig:seq_search_par_eval}    
\end{figure}

In Table~\ref{tab:shared_memory_learning_diff_eval}, we report SPILDL's sequential and parallel shared-memory learning (i.e. parallel search and parallel evaluation) on the 6 datasets, using different evaluation methods (provided by HT-HEDL). When the number of parallel search threads is 1, this means that the search runs sequentially (baseline hypothesis search). The 'sequential scalar CPU evaluation' refers to the baseline sequential (single core) hypothesis evaluation, which also is not accelerated by vector instructions. SPILDL extends upon the capabilities of DL-Learner's state of the art OCEL algorithm, with parallel computing to improve learning performance; the OCEL algorithm uses sequential hypothesis search and sequential hypothesis evaluation, for its DL-based ILP learning. In the experimental results, where both sequential hypothesis search and sequential hypothesis evaluation are used, these experiments reflect the learning performance of the OCEL algorithm; we use these experiments that reflect OCEL's learning performance as the baseline for SPILDL experiments -- because when both sequential search and sequential evaluation are used with SPILDL, this makes SPILDL identical to OCEL. See table~\ref{tab:seq_learning_diff_evals} for the summary  of experimental results on learning, using different evaluation methods. Also see Fig~\ref{fig:seq_search_par_eval} for a visualization on learning performance using different evaluations.

\begin{table}[h]
\caption{A summary for SPILDL's sequential learning using different evaluations.}
\label{tab:seq_learning_diff_evals}
\begin{tabularx}{\columnwidth}{X|X|X|X|X|X}
\hline
\multirow{2}{*}{\textbf{Dataset}} & \multicolumn{5}{c}{\textbf{Sequential search with different evaluations}} \\ \cline{2-6}
 & \textbf{Baseline} & \textbf{MT (scalar)} & \textbf{MT (vector)} & \textbf{GPU\textsubscript{1}} & \textbf{3 GPUs}\newline+\textbf{CPU} \\ \hline
Michalski trains & 126 & 148 & 148 & 154 & 158 \\ \hline
Moral & 1544 & 1609 & 1518 & 1492 & 1542\\ \hline
Carcino-genesis & 1791 & 1511 & 1428 & 1362 & 1386\\ \hline
IMDB & 10608 & 1861 & 704 & 472 & 279\\ \hline
Dunn-humby Retail & 125032 & 53576 & 44887 & 48900 & 50239\\ \hline
IMDB (string version) & 109117 & 26371 & 10706 & 8139 & 8966\\ \hline
\end{tabularx}
\end{table}
\FloatBarrier

In Fig~\ref{fig:seq_search_par_eval}, we can observe that each evaluation method has provided a speedup when compared to baseline performance (i.e. sequential evaluation). However, on smaller datasets (e.g. Michalski’s trains and Moral), parallel methods provide less performance than the baseline. The larger a dataset is, the more speedups are realized by parallel evaluations. In terms of CPU-based evaluations (i.e. scalar and vector), we can see that pure scalar multi-threading improves evaluation performance; we have also observed that multi-threaded performance is further increased (or amplified) when vector instructions (of each CPU core) are combined with multi-threading. 
In terms of a single GPU evaluations, the GPU outperformed the vectorized multi-threaded CPU evaluation on most datasets. In terms of multi-device evaluation, we can see that combining GPUs and CPUs results in performance gains outperforming other evaluation methods in some cases (e.g. IMDB dataset), or in other cases, provide performance gains comparable to single GPU and vectorized CPU evaluation methods. The multi-device evaluation method focuses on accelerating the evaluation when evaluating large number of hypotheses at once, by dividing the evaluation task among available processors (GPUs and CPUs). The multi-device evaluation method is most effective for ILP learning tasks where large number of hypotheses are generated in a single iteration, where the evaluation of single hypothesis is done against a dataset, that have large number of learning examples and ABox assertions.

\FloatBarrier

\begin{figure}[h]
\includegraphics[width=0.50\textwidth]{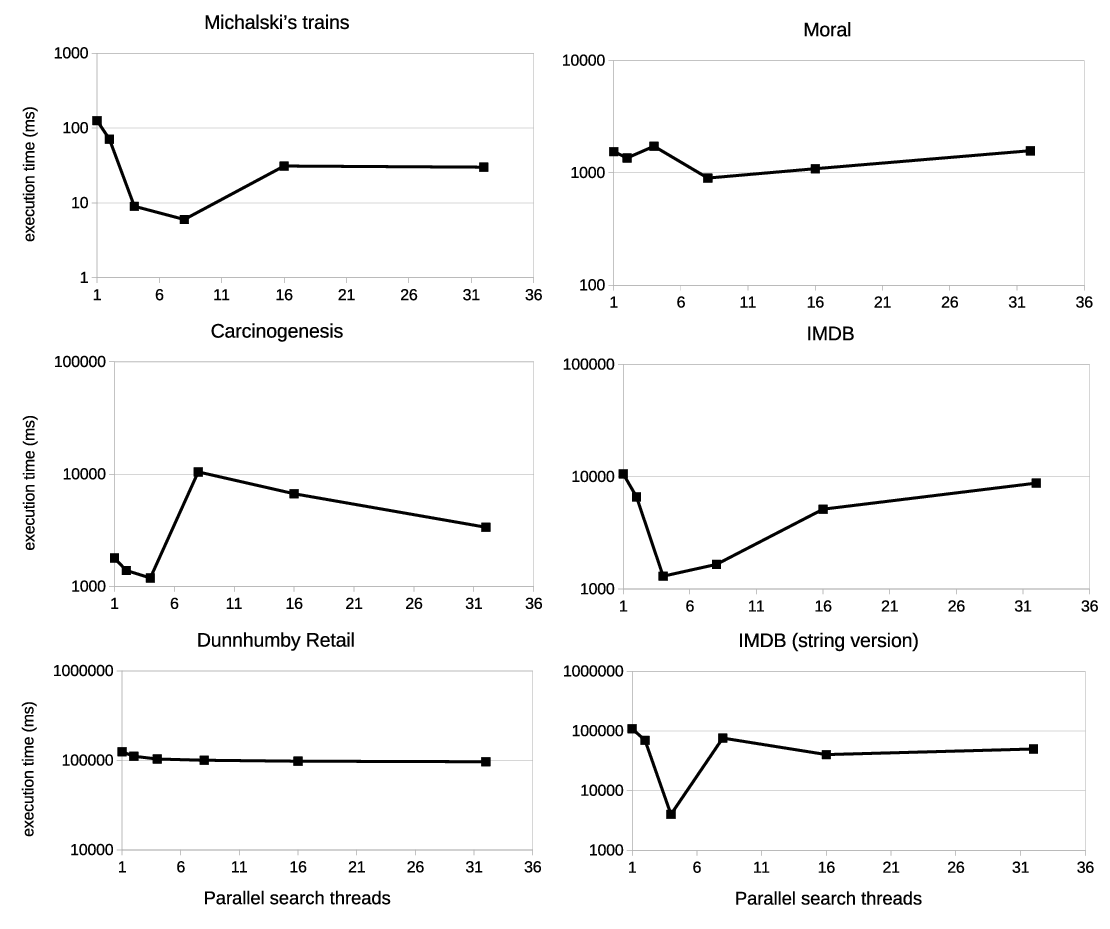}
\caption{The effect of parallel search on learning performance (using seq. evaluation).}
\label{fig:par_search_eval}    
\end{figure}

In terms of parallel hypothesis search, see Table~\ref{tab:par_learning_diff_eval} for a summary on parallel search experiments, and Fig~\ref{fig:par_search_eval} for visualization on these experiments. In Fig~\ref{fig:par_search_eval}, we observe that performance gains achieved by parallel search are dependent on the nature of each dataset -- that is, the nature of a dataset's search space. It is clear, that increasing the number of parallel search threads have improved learning performance on all datasets; however, the performance gains achieved by the parallel search grows up to a certain limit (i.e. the maximum number of parallel search threads), which changes depending on the dataset's search space.
In other words, choosing the number of parallel search threads which yields the maximum performance gains for the learning task, is dependent on the dataset's search space. However, we have observed that having 4 or 8 parallel search threads, provides the most performance gains on almost all datasets. See Table~\ref{tab:par_learning_diff_eval} for a summary on SPILDL's shared-memory learning experiments. 

\FloatBarrier

\begin{table}[!htb]
\caption{A summary for SPILDL's learning using parallel search.}
\label{tab:par_learning_diff_eval}
\begin{tabularx}{\columnwidth}{X|X|X|X|X|X}
\hline
\textbf{Dataset} & \multicolumn{5}{c}{\textbf{Parallel search with different evaluations}} \\ \cline{2-6}
\textbf{(\#search threads)} & \textbf{Baseline} & \textbf{MT (scalar)} & \textbf{MT (vector)} & \textbf{GPU\textsubscript{1}} & \textbf{3 GPUs}\newline+\textbf{CPU} \\ \hline
Michalski trains (8) & 6 & 7 & 7 & 8 & 8\\ \hline
Moral (8) & 897 & 927 & 907 & 905 & 908\\ \hline
Carcino-genesis (4) & 1190 & 872 & 808 & 774 & 769\\ \hline
IMDB (4) & 1302 & 279 & 95 & 60 & 41\\ \hline
Dunn-humby Retail (32) & 96617 & 25598 & 16730 & 21174 & 22938\\ \hline
IMDB (string version) (4) & 4002 & 993 & 370 & 256 & 195\\ \hline
\end{tabularx}
\end{table}
\FloatBarrier

\begin{figure}[!htb]
\includegraphics[width=0.48\textwidth]{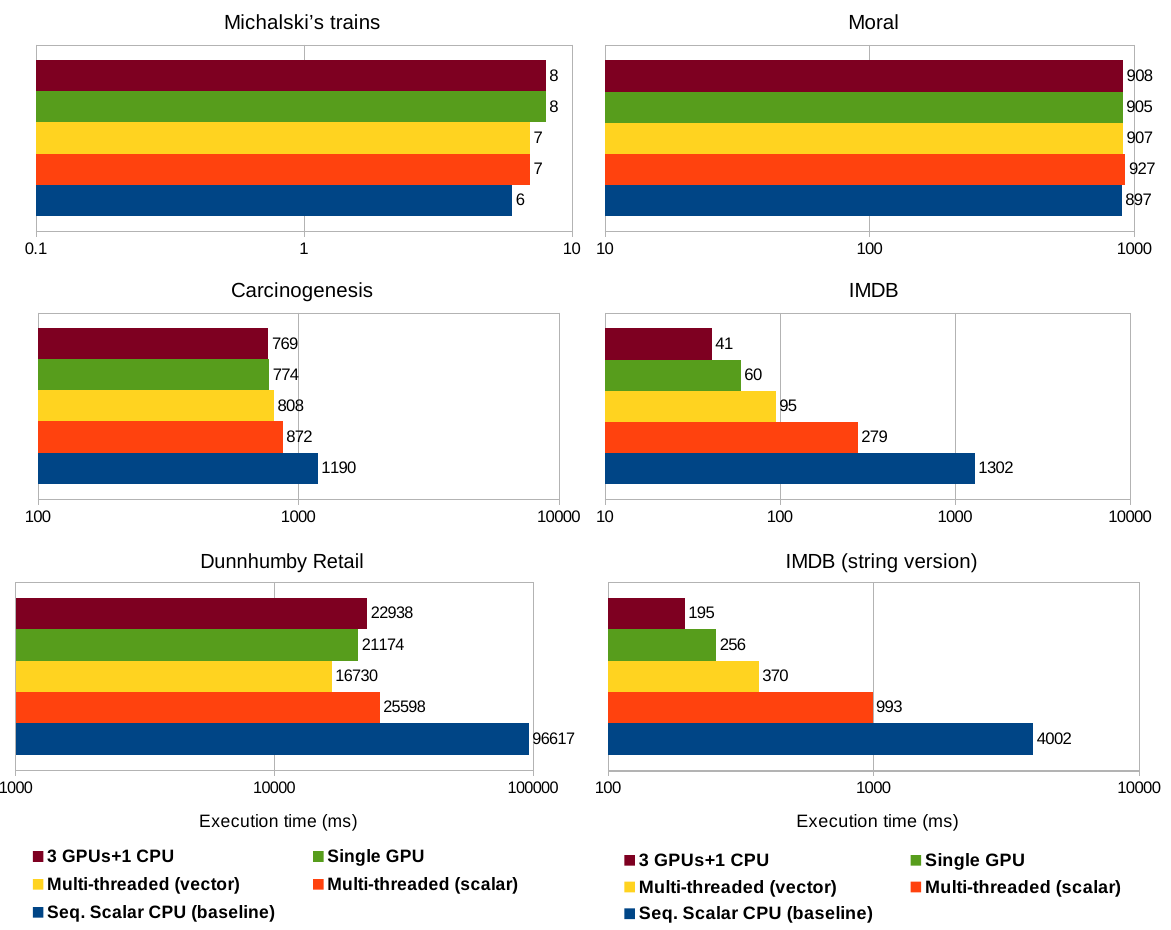}
\caption{A visualization for SPILDL's learning using parallel search.}
\label{fig:par_search_par_eval}    
\end{figure}
\FloatBarrier

In Table~\ref{tab:par_learning_diff_eval}, the numbers in parenthesis with each dataset, is the number of parallel search threads which yields the maximum search performance for that dataset. For a visualization on shared-memory experiments, see Fig~\ref{fig:par_search_par_eval}.
According to the summary of experimental results (in Table~\ref{tab:par_learning_diff_eval}), it is clear that parallel search and parallel evaluation improves learning performance with varying degrees of speedups -- these speedups are further amplified, when both parallel search and evaluation are combined.

\subsection{The experiments for cluster-based learning}
In the previous experiments, we have extensively evaluated the combinations of parallel search and parallel evaluation on all datasets. In this section, we provide experimental results on SPILDL's cluster-based learning. The main difference between shared-memory and cluster-based learning in SPILDL, is the locality of the CPUs that expands hypotheses. In shared-memory we expand hypotheses using the CPU cores within the same machine, whereas in cluster-based learning, we use TCP/IP networks to link (and communicate with) CPUs of networked machines. In terms of ILP learning, both learning modes (shared-memory \& cluster) have the same ILP learning characteristic (as a machine learning algorithm). However, the performance difference (in terms of learning speed) varies, depending on the locality of the CPUs (local vs remote CPUs).

Since both learning methods have identical machine learning performance, we evaluate cluster-based learning using 2 smallest and 2 largest datasets; where each dataset varies in the size of its search space and its ABox size. See Table~\ref{tab:remote_learning_exps} for a comparison between share-memory learning and cluster-based learning, using sequential search and with parallel evaluation; we have extensively studied the effect of different combinations of parallel search and parallel evaluation in the previous section. In this section (cluster-based learning), we are concerned with studying the effect of communication overhead on learning performance.

\begin{table}[!htb]
\caption{A comparison between SPILDL's local and cluster-based learning.}
\label{tab:remote_learning_exps}
\begin{tabularx}{\columnwidth}{p{1cm}|X|p{1.8cm}|X}
\hline
\multirow{2}{*}{\textbf{Dataset}} &\multicolumn{2}{c|}{\textbf{M\textsubscript{2} Shared-memory learning}} & \textbf{Cluster-based learning}\\ \cline{2-3}
 & \textbf{Sequential learning} & \textbf{Parallel evaluation learning} & (Master: \textbf{M\textsubscript{2}},\newline Worker: \textbf{M\textsubscript{1}})\\ \hline
Michalski trains & 240 & 251 & 448\\ \hline
Moral & 2353 & 2301 & 1569\\ \hline
IMDB & 12240 & 2343 & 367\\ \hline
Dunn-humby Retail & 192489 & 138840 & 80540\\ \hline
\end{tabularx}
\end{table}\FloatBarrier

\begin{figure}[!htb]
\includegraphics[width=0.49\textwidth]{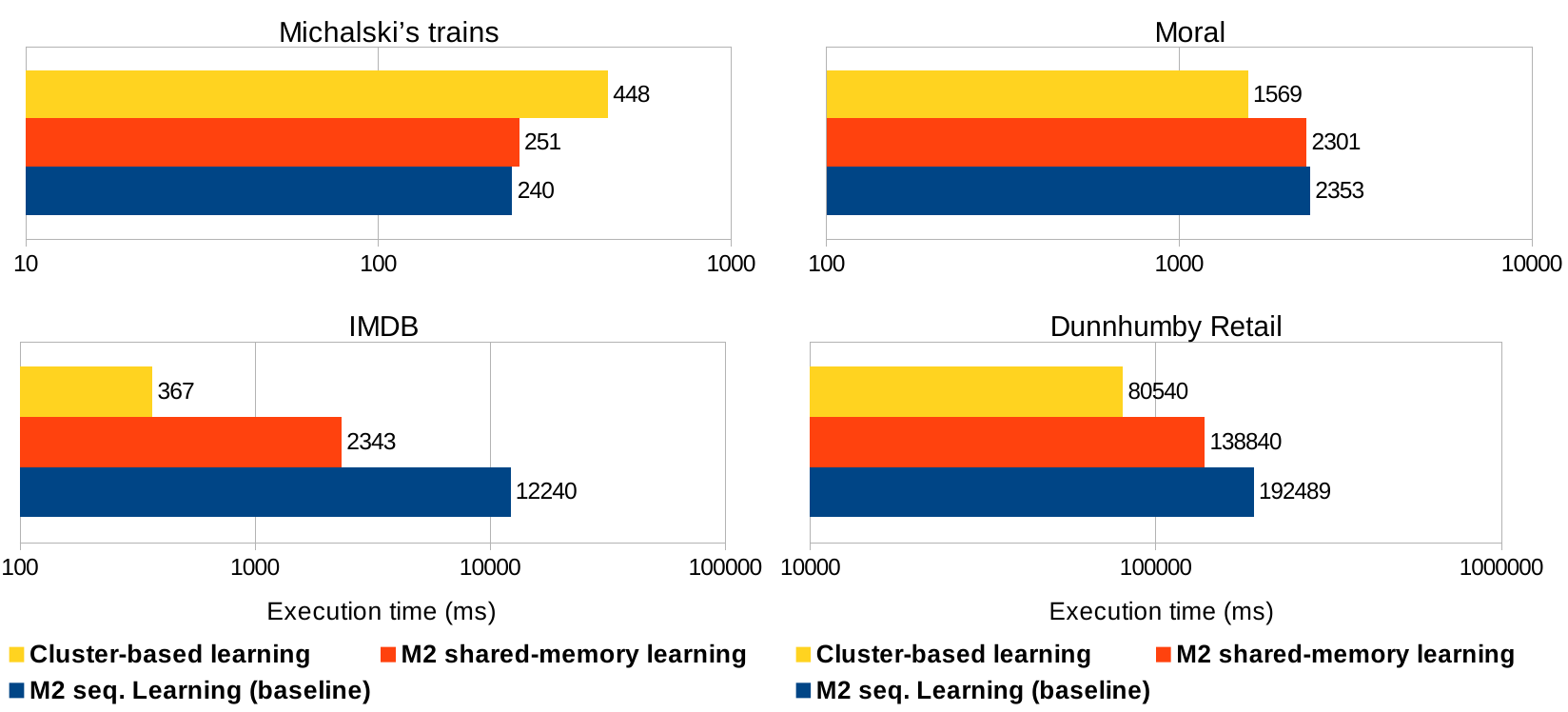}
\caption{A visualization of SPILDL's cluster-based learning.}
\label{fig:remote_learning}
\end{figure}

In Table~\ref{tab:remote_learning_exps}, the 4 datasets provides the performance boundaries that restricts the cluster-based learning; that is, the worst case scenario for cluster-based learning is reflected by the experiments on Michalski’s trains dataset, i.e. the smallest dataset which has the smallest search space and ABox size. However, the best case  scenario (for cluster-based learning) is reflected by the IMDB dataset, which has the largest search space and ABox size. See Fig~\ref{fig:remote_learning}, for a visualization on experimental results using cluster-based learning. 

\FloatBarrier

In Fig~\ref{fig:remote_learning}, we can observe that cluster-based learning will improve performance over baseline, when the dataset is large enough in terms of its search space and ABox size; with which communication overheads are canceled out (or at least justified) by performance gains achieved by  worker machine (s).

Based on the experimental results, we can observe that it is indeed true that parallel computing techniques do improve the performance of both hypothesis evaluation and hypothesis search tasks for DL-based ILPs. However, certain considerations has to be made regarding the datasets that will be used for DL-based ILP learning, in order for parallel computing techniques to achieve performance gains for DL-based ILPs. First, on small ILP datasets (e.g. Michalski trains), using either parallel search or parallel evaluation will provide performance similar (or worse) than baseline; because of parallel computing related overheads, which are: 
\begin{enumerate}
\item the overheads related to preparing, running, and terminating the parallel CPU threads (used for parallel search and/or parallel evaluation),
\item the overheads related to performing and also the handling of intermediate results for parallel redundancy checks (in Fig.~\ref{fig:redCheck}),
\item the overheads related to CPU-GPU communication -- for GPU-based evaluation, and
\item the overheads related to CPU+ multi-GPUs communication and workload distribution and scheduling -- for combined CPU+GPU evaluation,
\end{enumerate}

will cancel out any potential performance gains. Therefore, on small datasets, baseline methods (i.e. sequential search and evaluation) are sufficient to achieve reasonable ILP learning performance. Similar to Michalski trains dataset, on the Moral dataset, parallel evaluation provided either similar or worse performance than baseline.

Unlike small dataset, DL-based ILP learning from larger datasets such as the two IMDB datasets, and Dunnhumby dataset; has higher ILP learning performance through both parallel search and parallel evaluation approaches. In terms of parallel hypothesis evaluation, since evaluating hypotheses against large number of individuals is more computationally intensive than evaluating hypotheses against small number of individuals; therefore the performance gains achieved through parallel computing techniques, are high enough that they cancel out the aforementioned overheads -- related to the use of parallel computing techniques. In terms of parallel search, large ILP datasets has benefited the most from parallel search; because performing parallel search on large ILP datasets, results in avoiding (or minimizing) the generation and evaluation of large number of computationally expensive hypotheses, that may also belong to unpromising areas of the hypothesis search space. However, on small datasets the benefits of parallel search are canceled out by the related parallel computing overheads; because on small datasets, evaluating hypotheses is computationally cheap (relative to hypotheses on large datasets), and the cost of performing the parallel search is higher than the cost of evaluating these computationally cheap hypotheses -- we can see this effect by observing the parallel search experiments on the Moral dataset. Even though the Moral dataset has a considerably large search space, yet parallel search provided a performance similar or worse than baseline.

The performance of parallel search is sensitive to the nature of the search space for the used ILP dataset, and also to the number of CPU threads used to perform the parallel search. In other words, using higher number of CPU threads for parallel search, doesn't always translate into higher search performance -- we can observe this effect on most parallel search experiments. Although, each dataset has its own appropriate number of parallel search threads, that suits the nature of its search space. The reason for why higher number parallel threads, doesn't necessarily lead into higher search performance is because of the following reason. Higher number of parallel search threads lead into generating higher number of hypotheses in a single learning iteration -- which improve performance for some datasets such as Dunnhumby dataset. Some of these generated hypotheses may have high evaluation scores, which consequently make the learning algorithm choose them next for hypothesis expansion. Even though these generated hypotheses do have high evaluation scores, yet they may lead the learning algorithm into spending more time in areas of the search space that may seem initially promising, though may not contain solution hypotheses; as a result, this distract the learning algorithm, which negatively affect the parallel search performance.

In terms of SPILDL's limitations, there are limitations related shared-memory learning, and limitations related to cluster-based learning. In shared-memory learning, parallel computing performance is at maximum, because it has the least parallel computing overhead in comparison to cluster-based learning -- given the same hardware. However, in terms of scalability, the performance of shared-memory learning is constrained by the limit of which the computing and storage capabilities can be upgraded for the given machine. In other words, shared-memory learning achieves performance through vertical scalability, which is inherently limited by the upgrade capacity of the used machine. On the other hand, cluster-based learning offer better scalability, because when more computing power is needed, more machines can be added into the cluster -- which is a form horizontal scalability. However, cluster-based learning suffer from high communication and work distribution overheads (in comparison to shared-memory learning); because the network (typically TCP/IP) is used for communication. Although, similar to shared-memory learning, cluster-based learning also provide higher learning performance when the ILP dataset is computationally expensive enough, that it justifies the associated overheads; we can observe this effect on cluster-based learning experiments on Moral, IMDB, and Dunnhumby datasets. 

Even though SPILDL employ a hybrid scalability approach that combine both vertical and horizontal scaling for its cluster-based learning, yet, shared-memory learning has higher efficiency at using the parallel computing capabilities of available hardware. In other words, assuming a single very powerful machine that has an identical computing power equal to the aggregated computing power of a cluster of multiple weaker machines; in that case, the single very powerful machine will provide the highest ILP learning performance -- because the parallel computing overheads are at minimum. Moreover, upgrading the computing capabilities of a single machine such as adding more RAM or replacing the CPU with a more powerful CPU, is certainly cheaper than adding another machine with similar (or identical) computing power. However, cluster-based learning is preferred when learning from very large datasets -- especially when the very large datasets have also very large search spaces.

\section{Conclusion and future work}
Scalable machine learning is an important capability, that helps constructing powerful AI models in many real-world applications. However, constructing AI models from real-world data is a very challenging task - especially for ILPs-, since in many domains (such as e-commerce), the data is inherently large in size (GBs, or even TBs). 

In the context of this work, we have proposed different parallel approaches to accelerate ILP learning in description logic, which reduced -according to experimental results- the learning time, in addition to the ability of directly handling (real-world) large datasets; our parallel approaches mainly targets hypothesis search and hypothesis evaluation. For hypothesis search, we have provided two parallel search approaches, which accelerates the search using multi-core processors of: a single machine (i.e. shared-memory learning), and multiple networked machines in a cluster. 
In terms of hypothesis evaluation, we have combined (through HT-HEDL) the aggregated computing power of all multi-core CPUs and GPUs of single machine, to accelerate the evaluation task: for a single hypothesis, and for multiple hypotheses (i.e. batch of hypotheses). All the aforementioned parallel search approaches, uses HT-HEDL on each machine to utilize maximum evaluation performance on that machine.

According to experimental results, parallel search approaches improve performance by up to $\sim$27.3 folds, and parallel evaluation approaches (using HT-HEDL) improves performance by up to 38 folds. When both approaches are combined (i.e. parallel search and parallel evaluation), a speedup of up to $\sim$560 folds is achieved. For cluster-based learning, speedups of up to $\sim$33.4 folds were achieved (using only parallel hypothesis evaluation). In the worst case scenario, using parallel search doesn't always translate into higher search performance. In fact, on some ILP datasets, parallel search introduced performance gains similar or worse than baseline, while on other datasets, parallel search introduced higher performance gains faster than baseline; the parallel search performance is highly affected by the number of parallel search threads, the search space of the ILP dataset. In some ILP datasets, the baseline (sequential) search, provide the best search performance.
In terms of worst case scenario for parallel evaluation, the use of parallel evaluation (through HT-HEDL) result in similar or worse performance than baseline for small datasets; because on small datasets, parallel computing overheads are high enough that they cancel out any gained speedups. On the other hand, performance gains are consistently achieved for large datasets, where the achieved speedups vary depending on the used evaluation method.

For future directions, in terms of machine learning capabilities (as an ILP learner), SPILDL learning can be extended to include more expressive logics such as first-order-logic (Horn clauses in particular). In addition, we can develop case studies using SPILDL to learn multi-relational models describing variety of real-world concepts.

\section{Acknowledgment}
Special thanks to my PhD supervisor (Dr. Tommy Yuan) from the University of York for his support which helped this research.

\begin{IEEEbiographynophoto}{Eyad Algahtani} received his PhD degree in computer science with focus on scalable machine learning from the university of York, United Kingdom. In his studies, he is working on developing scalable Inductive Logic Programming (ILP) algorithms, capable of constructing human-interpretable Machine Learning (ML) models, from large amount of real-world data; through both HPC (High-performance computing) and non-HPC approaches. He is currently working as an assistant professor in King Saud University, Saudi Arabia. His main research interests focus on scalable and high-performance human-interpretable ML.
\end{IEEEbiographynophoto}

\vfill

\end{document}